\pdfoutput=1
\documentclass[11pt]{article}

\usepackage[]{acl}

\usepackage{times}
\usepackage{latexsym}
\usepackage{subcaption}
\usepackage{graphicx}
\usepackage{amsmath}
\usepackage{amssymb}
\usepackage{multirow}
\usepackage{algorithm}
\usepackage{algorithmic}
\usepackage{makecell}

\usepackage[T1]{fontenc}
\usepackage[utf8]{inputenc}

\usepackage{microtype}

\title{CLMLF:A Contrastive Learning and Multi-Layer Fusion Method for Multimodal Sentiment Detection}

\author{Zhen Li, Bing Xu\thanks{$^*$ Corresponding author}, Conghui Zhu, Tiejun Zhao \\
        Harbin Institute of Technology, Harbin, China \\
        \texttt{linklizhen@163.com}, \texttt{\{hitxb,conghui,tjzhao\}@hit.edu.cn}}

\begin{document}
\maketitle
\begin{abstract}

Compared with unimodal data, multimodal data can provide more features to help the model analyze the sentiment of data.
Previous research works rarely consider token-level feature fusion, and few works explore learning the common features related to sentiment in multimodal data to help the model fuse multimodal features.
In this paper, we propose a Contrastive Learning and Multi-Layer Fusion (CLMLF) method for multimodal sentiment detection.
Specifically, we first encode text and image to obtain hidden representations, and then use a multi-layer fusion module to align and fuse the token-level features of text and image.
In addition to the sentiment analysis task, we also designed two contrastive learning tasks, label based contrastive learning and data based contrastive learning tasks, which will help the model learn common features related to sentiment in multimodal data.
Extensive experiments conducted on three publicly available multimodal datasets demonstrate the effectiveness of our approach for multimodal sentiment detection compared with existing methods. The codes are available for use at \url{https://github.com/Link-Li/CLMLF}

\end{abstract}

\section{Introduction}

With the development of social networking platforms which have become the main platform for people to share their personal opinions. 
How to extract and analyze sentiments in social media data efficiently and correctly has broad applications. Therefore, it has attracted attention from both academic and industrial communities~\citep{zhang2018deep,yue2019survey}.
At the same time, with the increasing use of mobile internet and smartphones, more and more users are willing to post multimodal data (e.g., text, image, and video) about different topics to convey their feelings and sentiments. So multimodal sentiment analysis has become a popular research topic~\citep{kaur2019multimodal}.

\begin{figure}[tb]
\centering
\subfloat[Heathrow. Fly early tomorrow morning. (positive)]{
\includegraphics[width=0.21\textwidth]{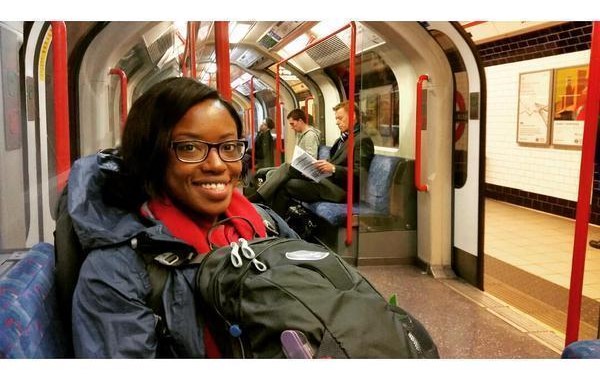}
\label{fig:case_example_fig_a}
}
\quad
\subfloat[Blue Jays game with the fam! Let's go! (positive)]{%
\includegraphics[width=0.21\textwidth]{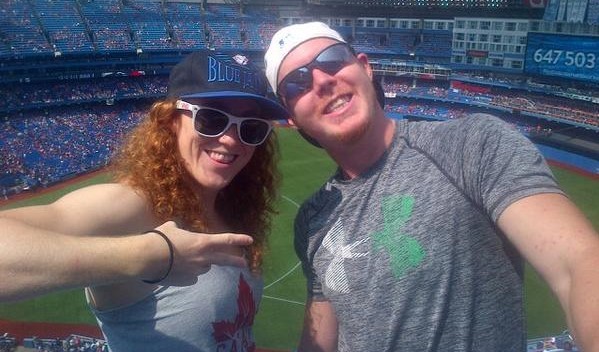}
\label{fig:case_example_fig_b}
}
\quad
\subfloat[Ridge Avenue is closed after a partial building collapse and electrical fire Saturday night. (negative)]{
\includegraphics[width=0.21\textwidth]{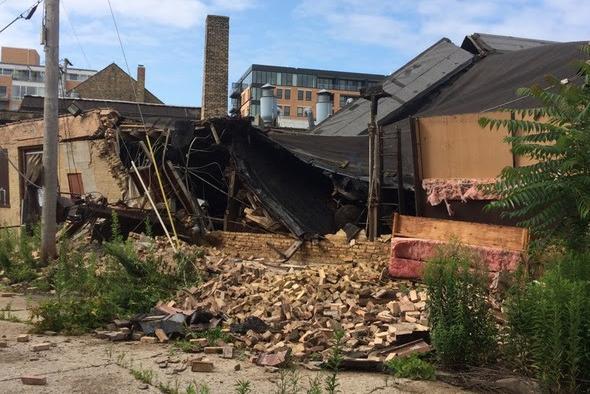}
\label{fig:case_example_fig_c}
}
\quad
\subfloat[Flexible spinal cord implants will let paralyzed people walk. (neutral)]{
\includegraphics[width=0.21\textwidth]{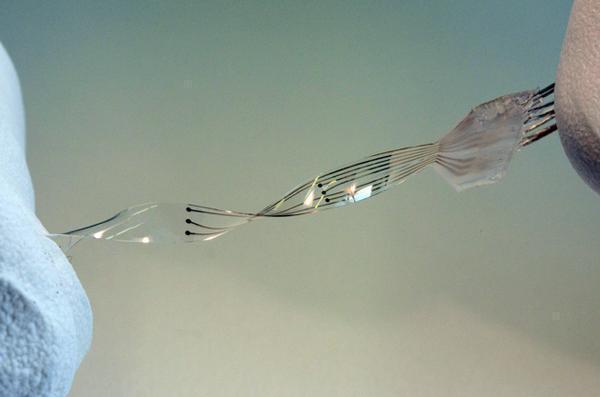}
\label{fig:case_example_fig_d}
}
\caption{Examples of multimodal sentiment tweets}
\label{fig:case_example_fig}
\end{figure}

As for multimodal data, the complementarity between text and image can help the model analyze the real sentiment of the multimodal data.
As shown in Figure~\ref{fig:case_example_fig}, detecting sentiment with only text modality or image modality may not be certain of the true intention of the tweet.
Such as Figure~\ref{fig:case_example_fig_a}, if we only analyze the text modality, we will find that this is a declarative sentence that does not express sentiment.
In fact, the girl's smile in the image shows that the sentiment of this tweet is positive.
At the same time, in Figure~\ref{fig:case_example_fig_c}, we can find that the ruins in the image which deepen the expression of negative sentiment in the text.

For multimodal sentiment analysis, we focus on text-image sentiment analysis in social media data.
In existing works, some models try to concatenate different modal feature vectors to fuse the multimodal features, such as MultiSentiNet~\citep{xu2017multisentinet} and HSAN~\citep{xu2017analyzing}.
\citet{kumar2020gated} proposes to use gating mechanism and attention to obtain deep multimodal contextual feature vectors.
Multi-view Attentional Network (MVAN) is proposed by \citet{yang2020image} which introduces memory networks to realize the interaction between modalities.
Although the above mentioned models are relatively better than unimodal models, the inputs with different modalities are in different vector spaces.
Therefore, it is difficult to fuse multimodal data with a simple concatenation strategy, so the improvement is also limited.
Furthermore, the gating mechanism and memory network are essentially not designed for multimodal fusion.
Although they can help the model analyzes the sentiment in the multimodal data by storing and filtering the features in the data, it is obvious that these methods are difficult to align and fuse the features of text and image.
Since Transformers have achieved great success in many fields, such as natural language processing and computer vision~\citep{lin2021survey, khan2021transformers}, we propose \textbf{M}ulti-\textbf{L}ayer \textbf{F}usion (\textbf{MLF}) module based on Transformer-Encoder. Benefiting from the multi-headed self-attention in Transformer, which can capture the internal correlation of data vectors.
Therefore, text tokens and image patches with explicit and implicit relationships will have higher attention weight allocation to each other which means the MLF module can help align and fuse the token-level text and image features better.
And MLF is a multi-layer encoder, which can help improve the abstraction ability of the model and obtain deep features in multimodal data.

Some previous work has explored the application of contrastive learning in the multimodal field. \newcite{huang2021multilingual} proposes the application of contrastive learning in multilingual text-to-video search, and \newcite{yuan2021multimodal} applies contrastive learning to learn visual representations that embraces multimodal data. However, there is little work to study the application of contrastive learning in multimodal sentiment analysis,  so we propose two contrastive learning tasks, \textbf{L}abel \textbf{B}ased \textbf{C}ontrastive \textbf{L}earning (\textbf{LBCL}) and \textbf{D}ata \textbf{B}ased \textbf{C}ontrastive \textbf{L}earning (\textbf{DBCL}), which will help the model learn common features related to sentiment in multimodal data. For example, as shown in Figure~\ref{fig:case_example_fig_a} and Figure~\ref{fig:case_example_fig_b}. We can find that both tweets show positive sentiment. And we also can find there are smiling expressions in the image of the two tweets which is a common feature of those tweets. If the model can learn common features related to sentiment, it will greatly improve the performance of the model.

In this paper, we propose a \textbf{C}ontrastive \textbf{L}earning and \textbf{M}ulti-\textbf{L}ayer \textbf{F}usion (\textbf{CLMLF}) method for multimodal sentiment analysis based on text and image modalities.
For evaluation, CLMLF is verified on three multimodal sentiment datasets, namely MVSA-Single, MVSA-Multiple~\citep{niu2016sentiment} and HFM~\citep{cai2019multi}.
CLMLF achieves better performance compared to several baseline models in all three datasets.
Through a comprehensive set of ablation experiments, case study, and visualizations, we demonstrate the advantages of CLMLF for multimodal fusion\footnote{There are also the experimental results and analysis of CLMLF in aspect based multimodal sentiment analysis task, which can refer to Appendix~\ref{sec:aspect_based_multimodal_sentiment}}.
Our main contributions are summarized as follows:

\begin{itemize}

\item {We propose a multi-layer fusion module based on Transformer-Encoder that multi-headed self-attention can help align and fuse token-level features of text and image, and it can also benefit from the depth of MLF which improves model abstraction ability.
Experiments show that the proposed architecture of MLF is simple but effective.}

\item {We propose two contrastive learning tasks based on label and data, which leverages sentiment label features and data augmentation.
Those two contrastive learning tasks can help the model learn common features related to sentiment in multimodal data, which improve the performance of the model.}

\end{itemize}

\section{Approach}

\subsection{Overview}

\begin{figure*}[htbp]
    \centering
    \centering\includegraphics[width=0.9\textwidth]{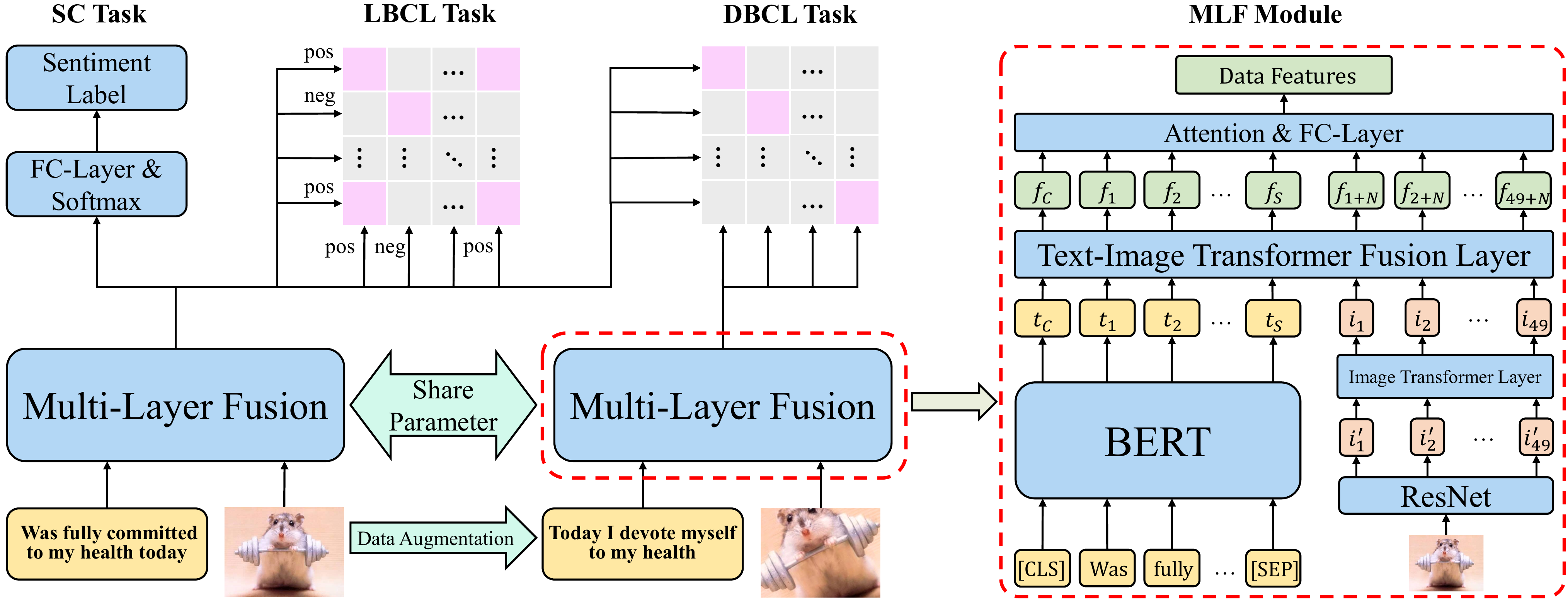}
    \caption{The framework of the proposed CLMLF model}
    \label{fig:CLMLF}
\end{figure*}

In this section, we will introduce CLMLF. Figure~\ref{fig:CLMLF} illustrates the overall architecture of CLMLF model for multimodal sentiment detection that consists of two modules: multi-layer fusion module and multi-task learning module. Specifically, the multi-layer fusion module is on the right in Figure~\ref{fig:CLMLF}, it includes a text-image encoder, image Transformer layer, and text-image Transformer fusion layer modules. The multi-task learning module is on the left in Figure~\ref{fig:CLMLF},  it includes three tasks, sentiment classification, label based contrastive learning and data based contrastive learning tasks.

\subsection{Multi-Layer Fusion Module}

We use Multi-Layer Fusion module to align and fuse the token-level features of text and image. As shown on the right of Figure~\ref{fig:CLMLF}. First, we use BERT~\citep{devlin2018bert} and ResNet~\citep{he2015deep} to encode the text and image to obtain the hidden
representation of the text $T = \{ t_C, t_1, t_2, ..., t_S \}, T \in \mathbb{R}^{n_t \times d_t} $ and the hidden
representation of the image $I_c^{'} \in \mathbb{R}^ {p_i \times p_i \times d_i} $, and $I_c^{'}$ is the image feature map output by the last layer of convolution layer of ResNet. We transform the hidden
representation dimension of $I_c^{'}$ into the same dimension as the  $T$. And we can get the sequence feature representation of the image $I^{'}$ as follows: 

\begin{small}
\begin{gather}
    I^{'} = flatten(I_c^{'} W_{I} + b_{I}) 
\end{gather}
\end{small}
Where $I^{'} = \{ i_{1}^{'}, i_{2}^{'}, ..., i_{n_i}^{'} \}, I^{'} \in \mathbb{R}^{n_i \times d_t}$, $n_i=p_i \times p_i$. And the function of $flatten$ means flatten the input vector by reshaping the first two-dimensions into a one-dimensional.

After that, we will encode the image sequence features $I^{'}$. Here we use the vanilla Transformer-Encoder proposed by \newcite{vaswani2017attention}. Input $I^{'}$ into the image Transformer layer which is based on a multi-layer Transformer-Encoder to obtain the final encoding of image sequence features $I$.

\begin{small}
\begin{gather}
    \{ i_{1}, i_{2}, ..., i_{n_i} \} = TE_{I}(\{ i_{1}^{'}, i_{2}^{'}, ..., i_{n_i}^{'} \}) \\
    I = \{ i_{1}, i_{2}, ..., i_{n_i} \}
\end{gather}
\end{small}

Where $TE_{I}$ means the vanilla Transformer-Encoder of image.

In order to align and fuse the features of text and images, we concatenate the features of the text $T$ and the image sequence features $I$. We use a new multi-layer Transformer-Encoder as a text-image fusion layer which will align and fuse multimodal features. Then the fusion sequence features of text and image can be obtained. It is as follows:

\begin{small}
\begin{gather}
    \{ f_{1}, f_{2}, ..., f_{n_t + n_i} \} = TE_{M}(concat(T, I)) \\
    F = \{ f_{1}, f_{2}, ..., f_{n_t + n_i} \}
\end{gather}
\end{small}

Where $TE_{M}$ means the vanilla Transformer-Encoder of multimodal data.

Now, we obtain the sequence features of text and image fusion, but it is obvious that the sequence features can not be used in the classification task. So we use a simple attention layer to get the multimodal representation $R$.

\begin{small}
\begin{gather}
    \tilde{q}_i = GELU(f_i W_1 + b_1) W_2 + b_2  \\ 
    q_i = exp(\frac{\tilde{q}_i}{\sum_{j=1}^{n_t+n_i}\tilde{q}_j}) \\
    \tilde{R} = \sum_{i=1}^{n_t+n_i} q_i f_i \\
    R = GELU(\tilde{R} W_R + b_R)
\end{gather}
\end{small}
where $GELU$ is the activation function. $R \in \mathbb{R}^{d^t}$

\subsection{Sentiment Classification}

As shown in the SC task in Figure~\ref{fig:CLMLF}, we feed the above multimodal representation $R$ into the fully connected layer and employ the softmax function for sentiment detection. We use the cross-entropy loss as the classification loss and it is as follows:

\begin{small}
\begin{gather}
    L_{sc} = Cross\text{-}Entropy(GELU(R W_{sc} + b_{sc})) 
\end{gather}
\end{small}

\subsection{Label Based Contrastive Learning}

In order to let the model learn the sentiment related features in the multimodal data, we use label based contrastive learning task to help the model extract the sentiment related features while MLF module fuses text and image data. As shown in the LBCL task in Figure~\ref{fig:CLMLF}, we divide the data in each batch into positive and negative examples according to its sentiment label. For example, in Figure~\ref{fig:CLMLF}, for a negative label of multimodal data, the data in the batch with the same negative labels as positive examples (the square of pink color), and the data with no negative labels are taken as negative examples (the square of gray color). 

The specific step can refer to Algorithm~\ref{alg:label_cl}. The meanings of specific functions in the algorithm are as follows: $einsum$ means Einstein summation convention, $gather$ means gathers values along with an index, and $\tau$ represents the contrastive learning's temperature. The algorithm consists of two main steps: the first step is to generate the unmask label $L_t$ according to the data labels in the batch; the second step is to calculate the loss matrix $l_{pn}$, and use the unmask label $L_t$ and the loss matrix $l_{pn}$ to get the final loss $L_{l\text{-}cl}$, which are the water-red elements in LBCL task on the left in Figure~\ref{fig:CLMLF}.

\begin{algorithm}[ht]
\caption{LBCL Algorithm}
\label{alg:label_cl}
% \begin{algorithmic}[1]
\begin{algorithmic}[1]
% \fontsize{9}{10}\selectfont
\REQUIRE The sentiment label is $L$, which is a list of all data in the batch, assuming that the sentiment is divided into three categories: positive (0), neutral (1) and negative (2); The Multi-Layer Fusion Model of $MLF$; the texts are $T$; the images are $I$; $C$ denotes length of $L_c$; $S$ denotes length of $L$.
\ENSURE Label contrastive learning loss $L_{l\text{-}cl}$
\STATE initialize $L_c = [L-0, L-1, L-2]$ and $L_t=list()$
\FOR {$i=1;i<=C;i++$} 
    \STATE initialize $\tilde{L}_t=list()$ 
    \FOR {$j=1;j<=T;j++$}
        \IF {$L_c[i][j]$ equals 0}
            \STATE $\tilde{L}_t.append(j)$
        \ENDIF
    \ENDFOR
    \STATE $L_t.append(\tilde{L}_t)$
\ENDFOR
\STATE $R = MLF(T, I)$
\STATE $\tilde{l}_{pn} =  einsum(nc, ck->nk, [R, R^T])$
\STATE $l_{pn} = LogSoftmax(l_{pn} / \tau).view(-1)$
\STATE $L_{cl} = L_t[L[1]]$
\FOR {$q=2; q<=S, q++$}
    \STATE $L_{cl} =  concat(L_{cl}, L_t[L[q]] + q \times T)$
\ENDFOR
\STATE $L_{lbcl} = gather(l_{pn}, index=L_{cl}) / T$
\RETURN $L_{lbcl}$
\end{algorithmic}
\end{algorithm}

\subsection{Data Based Contrastive Learning}

In order to strengthen the robustness of the model to the data and enhance the learning ability of the model to the invariant features in the data. We add a contrastive learning task based on data augmentation which is DBCL task in Figure~\ref{fig:CLMLF}. 
Considering the flexible expression of text and images. It may cause the model to be too sensitive to the surface features of data, rather than focus on fusing the invariant features in text and images, that is, effective features. Sentiment related features should exist in these effective features, because the true meaning of the meaning user wants to express should not change with the changes in text and images. For example, both "I had ice cream today. I was very happy" and "I'm very happy today because I ate ice cream" express positive sentiment. The keyword "happy" has not changed which means the happy is an effective feature, but some other words have changed greatly. The data based contrastive learning can force the model learning the effective features in the data, which is more conducive to the model to learn the features related to sentiment in the data. Algorithm~\ref{alg:data_cl} describes the process of data based contrastive learning.

Specifically, as for text, we use a data augmentation method called back-translation~\citep{sennrich2015improving, edunov2018understanding, xie2019unsupervised}, which refers to the procedure of translating an existing text $x$ in language $E$ into another language $C$ and then translating it back into $E$ to obtain
an augmented text $x$. As observed by \citet{yu2018qanet}, back-translation can generate diverse paraphrases while preserving the semantics of the original sentences. So we use back-translation to construct positive examples of text in contrastive learning.

For image augmentation, we use a method called RandAugment~\citep{cubuk2019randaugment}, which is inspired by AutoAugment~\citep{cubuk2018autoaugment}. AutoAugment uses a search method to combine all transformations to find a good augmentation strategy. In RandAugment, it does not use search, but instead uniformly samples from the same set of augmentation transformations. In other words, RandAugment is simpler and requires no labeled data as there is no need to search for optimal policies.

\begin{algorithm}[ht]
\caption{DBCL Algorithm}
\label{alg:data_cl}
% \begin{algorithmic}[1]
\begin{algorithmic}[1]
% \fontsize{9}{10}\selectfont
\REQUIRE The Multi-Layer Fusion Model of $MLF$; the texts are $T$; the images are $I$; $BT$ means back-translation and $RA$ means RandAugment; $S$ denotes of batch size.
\ENSURE Data contrastive learning loss $L_{d\text{-}cl}$

\STATE $R = MLF(T, I)$ 
\STATE $R_{au} = MLF(BT(T), RA(I))$
\STATE $l_{pn} = einsum(nc,ck->nk, [R, R_{au}^T])$
\STATE $cl\text{\_}label = arange(S)$
\STATE $L_{dbcl} = Cross\text{-}Entropy(l_{pn} / \tau, cl\text{\_}label)$
\RETURN $L_{dbcl}$
\end{algorithmic}
\end{algorithm}

\subsection{Model Training}

The label contrastive loss or data contrastive loss can be simply added to the total loss as a regularization. Can be combined like follows:

\begin{small}
\begin{equation}
\begin{aligned}
    L & =  L_{sc} + \lambda_{lbcl} L_{lbcl} + \lambda_{dbcl} L_{dbcl}
\end{aligned}
\label{equ:all_loss}
\end{equation}
\end{small}
where $\lambda_{lbcl}$ and $\lambda_{dbcl}$ are coefficients to balance the different training losses.

\begin{table*}[htbp]
\centering
\begin{tabular}{c|c|cc|cc|c|cc}
\hline
\multirow{2}{*}{\textbf{Modality}} & \multirow{2}{*}{\textbf{Model}} & \multicolumn{2}{c|}{\textbf{MVSA-Single}} & \multicolumn{2}{c|}{\textbf{MVSA-Multiple}} & \multirow{2}{*}{\textbf{Model}} &  \multicolumn{2}{c}{\textbf{HFM}} \\
& & Acc & F1 & Acc & F1 & & Acc & F1 \\
\hline
\multirow{4}{*}{\textbf{Text}} & CNN & 0.6819 & 0.5590 & 0.6564 & 0.5766 & CNN & 0.8003 & 0.7532  \\
& BiLSTM & 0.7012 & 0.6506 & 0.6790 & 0.6790 & BiLSTM & 0.8190 & 0.7753\\
& BERT & 0.7111 & 0.6970 & 0.6759 & 0.6624 & BERT & 0.8389 & 0.8326\\
& TGNN & 0.7034 & 0.6594 & 0.6967 & 0.6180  &\\
\hline
\multirow{2}{*}{\textbf{Image}} 
& ResNet-50 & 0.6467 & 0.6155 & 0.6188 & 0.6098 & ResNet-50 & 0.7277 & 0.7138 \\
& OSDA & 0.6675 & 0.6651 & 0.6662 & 0.6623 & ResNet-101 & 0.7248 & 0.7122 \\  
\hline
\multirow{5}{*}{\textbf{Multimodal}} & MultiSentiNet & 0.6984 & 0.6984 & 0.6886 & 0.6811 & Concat(2) & 0.8103 & 0.7799\\
& HSAN & 0.6988 & 0.6690 & 0.6796 & 0.6776  & Concat(3) & 0.8174 & 0.7874\\   
& Co-MN-Hop6 & 0.7051 & 0.7001 & 0.6892 & 0.6883 & MMSD & 0.8344 & 0.8018\\  
& MGNNS & 0.7377 & 0.7270 & 0.7249 & 0.6934  & D\&R Net & 0.8402 & 0.8060\\  
& CLMLF & \textbf{0.7533} & \textbf{0.7346} & 0.7200 & \textbf{0.6983}  & CLMLF & \textbf{0.8543} & \textbf{0.8487} \\  
\hline
\end{tabular}

\caption{Experimental results of different models on MVSA-Single, MVSA-Multiple and HFM datasets}
\label{tab:result}
\end{table*}

\section{Experimental Setup}

\subsection{Dataset}

We demonstrate the effectiveness of our method on three public datasets which are MVSA-Single, MVSA-Multiple\footnote{http://mcrlab.net/research/mvsa-sentiment-analysis-on-multi-view-social-data/}~\citep{niu2016sentiment} and HFM\footnote{https://github.com/headacheboy/data-of-multimodal-sarcasm-detection}~\citep{cai2019multi}. Both datasets collect data from Twitter, each text-image pair is labeled by a single sentiment. For a fair comparison, we process the original two MVSA datasets in the same way used in \citet{xu2017multisentinet}, as for HFM, we adopt the same data preprocessing method as that of \citet{cai2019multi}. We randomly split the MVSA datasets into train set, validation set, and test set by using the split ratio 8:1:1. The statistics of these datasets are given in Table~\ref{tab:data_statistics}. The detailed statistics of these datasets are given in Appendix~\ref{sec:dataset_detailed_statistics}.

\begin{table}[htbp]
\centering
\begin{tabular}{c|cccc}
\hline
\textbf{Dataset} & \textbf{Train} &  \textbf{Val} & \textbf{Test} & \textbf{Total} \\
\hline
MVSA-S & 3611 & 450 & 450 & 4511 \\
MVSA-M & 13624 & 1700 & 1700 & 17024 \\
HFM & 19816 & 2410 & 2409 & 24635 \\
\hline
\end{tabular}

\caption{Statistics of the three datasets}
\label{tab:data_statistics}
\end{table}

\subsection{Implementation Details}

For the experiments of CLMLF, we use the Pytorch\footnote{https://pytorch.org/} and HuggingFace Transformers\footnote{https://github.com/huggingface/transformers}~\citep{wolf-etal-2020-transformers} as the implementation of baselines and our method. We use the Bert-base\footnote{https://huggingface.co/bert-base-uncased} and ResNet-50\footnote{https://pytorch.org/vision/stable/models.html}
% \footnote{https://download.pytorch.org/models/resnet50-0676ba61.pth} 
as the text and image encoder in Multi-Layer Fusion module. 
The batch size is set to 32, 64 and 128 for MVSA-Single, MVSA-Multiple and HFM. We use AdamW optimizer. The $\epsilon$ is 1e-8 and $\beta$ is (0.9, 0.999). The learning rate is set to 2e-5. Both $\lambda_{lbcl}$ and $\lambda_{dbcl}$ are set to 1.0 in Equation~\ref{equ:all_loss} during training. For the number of layers of MLF, please refer to Section~\ref{sec:influence_of_mlf_layer}. And all the experiments are done on four NVIDIA 3090 GPUs.

\subsection{Compared Methods}

We compare our model with the unimodal sentiment models and the multimodal baseline models.

\textbf{Unimodal Baselines: } For text modality, CNN~\citep{kim-2014-convolutional} and Bi-LSTM~\citep{zhou-etal-2016-attention} are well-known models for text classification tasks. TGNN~\citep{huang2019text} is a text-level graph neural network for text classification. BERT~\citep{devlin2018bert} is a pre-trained model for text, and we fine-tuned on the text only. For image modality, OSDA~\citep{yang2020image} is an image sentiment analysis model based on multiple views. ResNet~\citep{he2015deep} is pre-trained and fine-tuned on the image only.

\textbf{Multimodal Baselines: } MultiSentiNet~\citep{xu2017multisentinet} is a deep semantic network with attention for multimodal sentiment analysis. HSAN~\citep{xu2017analyzing} is a hierarchical semantic attentional network based on image captions for multimodal sentiment analysis. Co-MN-Hop6~\citep{xu2018co} is a co-memory network for iteratively modeling the interactions between multiple modalities. MGNNS~\citep{yang2021multimodal} is a multi-channel graph neural networks with sentiment-awareness for image-text sentiment detection. \citet{schifanella2016detecting} concatenates different feature vectors of different modalities as multimodal feature representation. Concat(2) means concatenating text features and image features, while Concat(3) has one more image attribute features. MMSD~\citep{cai2019multi} fuses text, image, and image attributes with a multimodal hierarchical fusion model. \citet{xu2020reasoning} proposes the D\&R Net to fuse text, image, and image attributes by constructing the Decomposition and Relation Network.

\section{Results and Analysis}
\label{sec:results_and_analysis}

\begin{table*}[htbp]
\centering
\begin{tabular}{ccc|cc|cc|cc}
\hline
 & \multirow{2}{*}{\textbf{Model}} && \multicolumn{2}{c|}{\textbf{MVSA-Single}} & \multicolumn{2}{c|}{\textbf{MVSA-Multiple}} & \multicolumn{2}{c}{\textbf{HFM}} \\
 &&& Acc & F1 & Acc & F1 & Acc & F1 \\
\hline
 & BERT && 0.7111 & 0.6970 & 0.6759 & 0.6624 & 0.8389 & 0.8326 \\
 & ResNet-50 && 0.6467 & 0.6155 & 0.6188 & 0.6098 & 0.7277 & 0.7138 \\
 & +MLF && 0.7111 & 0.7101 & 0.7059 & 0.6849 & 0.8414 & 0.8355  \\
 & +MLF, LBCL && 0.7378 & 0.7291 & 0.7112 & 0.6863 & 0.8489 & 0.8446  \\
 & +MLF, DBCL && 0.7356 & 0.7276 & 0.7153 & 0.6832 & 0.8468 & 0.8422  \\
\hline
 & CLMLF && \textbf{0.7533} & \textbf{0.7346} & \textbf{0.7200} & \textbf{0.6983} & \textbf{0.8543} & \textbf{0.8487}  \\
\hline
\end{tabular}

\caption{Ablation results of our CLMLF}
\label{tab:ablation}
\end{table*}

\subsection{Overall Result}

Table~\ref{tab:result} illustrates the performance comparison of our CLMLF model with the baseline methods. We use Weighted-F1 and ACC as the evaluation metrics for MVSA-Single and MVSA-Multiple which is the same as \newcite{yang2021multimodal} and use Macro-F1 and ACC as the evaluation metrics for HFM. we have the following observations. First of all, our model is competitive with the other strong baseline models on the three datasets. Second, the multimodal models perform better than the unimodal models on all three datasets. What is more, we found the sentiment analysis on the image modality gets the worst results, this may be that the sentimental features in the image is too sparse and noisy, which makes it difficult for the model to obtain effective features for sentiment analysis. At last, for simple tasks, the performance improvement of multimodal models is limited. For example, on HFM dataset, the improvement of CLMLF relative to BERT is less than MVSA-Single dataset that because HFM is a binary classification task, while MVSA-Single is a three classification task.

We also try to apply CLMLF to aspect based multimodal sentiment analysis task which can refer to Appendix~\ref{sec:aspect_based_multimodal_sentiment} for details.

\subsection{Ablation}

We further evaluate the influence of multi-layer fusion module, label based contrastive learning, and data based contrastive learning. The evaluation results are listed in Table~\ref{tab:ablation}. The Result shows that the whole CLMLF model achieves the best performance among all models. We can see multi-layer fusion module can improve the performance, which shows that a multi-layer fusion module can fuse the multimodal data. On this foundation, adding the label and data based contrastive learning can improve the model performance more, which means contrastive learning can lead the model to learn common features about sentiment and lead different sentiment data away from each other.

\subsection{Influence of MLF Layer}
\label{sec:influence_of_mlf_layer}

\begin{figure}[htbp]
\centering
\subfloat[The text-image Transformer fusion layer]{%
\includegraphics[width=0.21\textwidth]{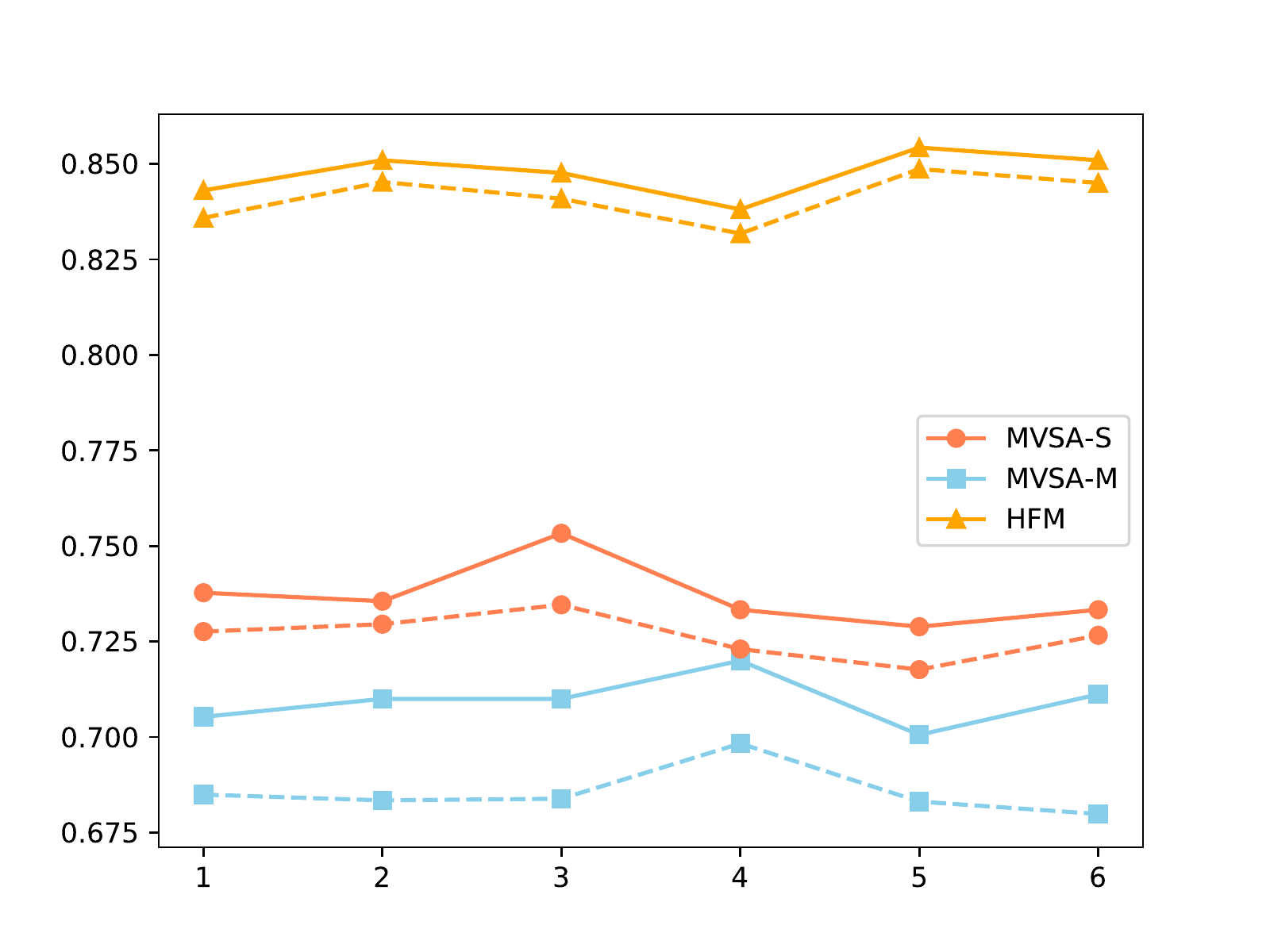}
\label{fig:MLF_multi_layer_result}
}
\quad
\subfloat[The image Transformer layer]{%
\includegraphics[width=0.21\textwidth]{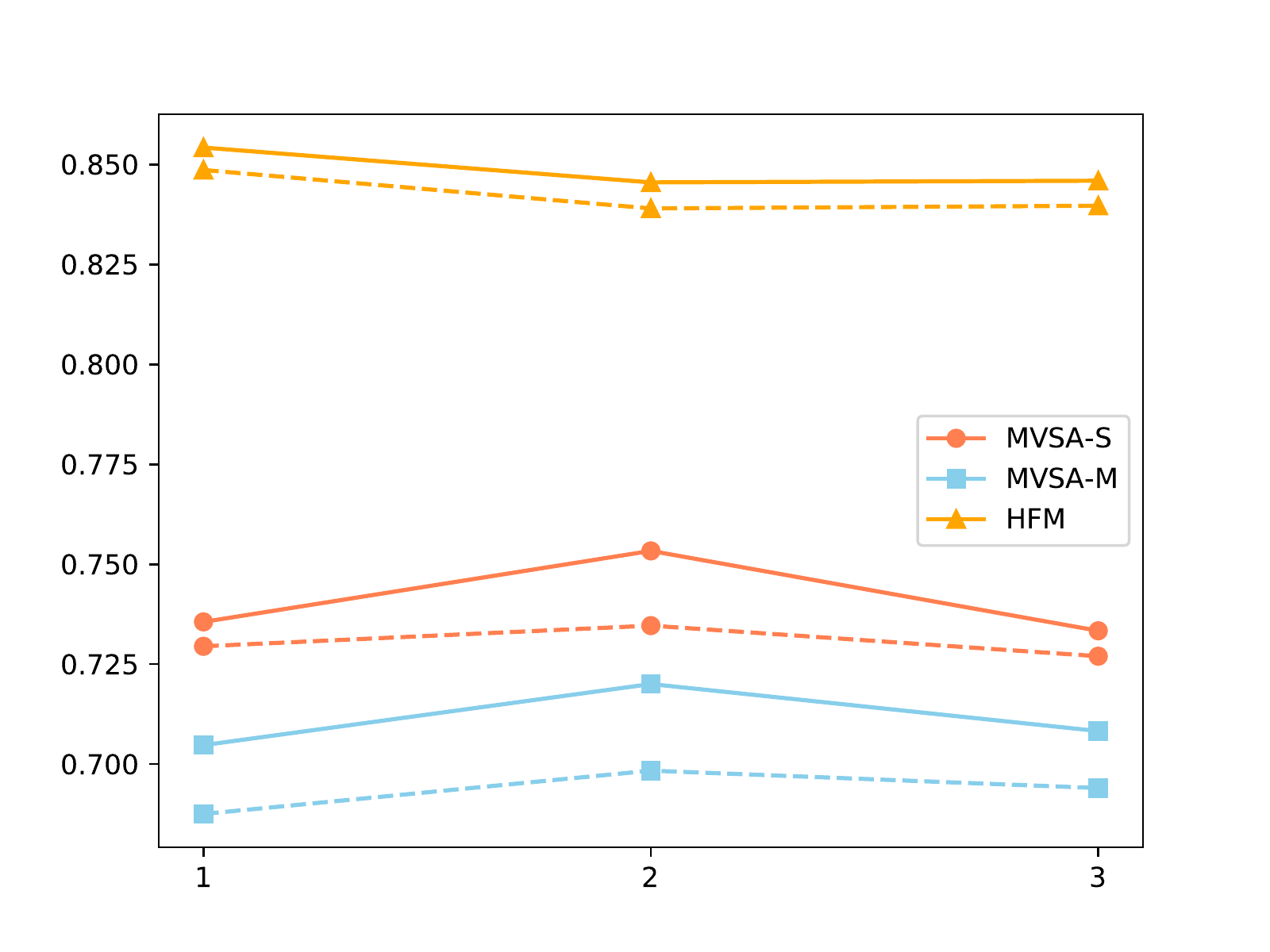}
\label{fig:MLF_image_layer_result}
}
\caption{Experimental results of different layer of multi-layer fusion module. The solid line indicates the accuracy and the dotted line indicates the F1. The x-axis represents the number of layers of the transformer}
\label{fig:MLF_layer_result}
\end{figure}

We explored the effects of different layers of Transformer-Encoder on the results. As shown in Figure~\ref{fig:MLF_multi_layer_result}, fix the image transformer layer and set the text-image transformer fusion layer from 1 to 6. As shown in Figure~\ref{fig:MLF_image_layer_result}, fix the text-image transformer fusion layer and set the image transformer layer from 1 to 3. Finally, we selected different combinations of 3-2 (which means three layers of text-image transformer fusion layer and two layers of image transformer layer), 4-2, and 5-1 for the three datasets. This also proves that the contribution of text and images in the dataset is different. It can be seen from Table~\ref{tab:result} that CLMLF gains more from the text than images in HFM dataset. Therefore, in MLF module, the layers of transformer related to text are more than images.

\subsection{Case Study}

\begin{figure}[htbp]
    \centering
    \centering\includegraphics[width=0.49\textwidth]{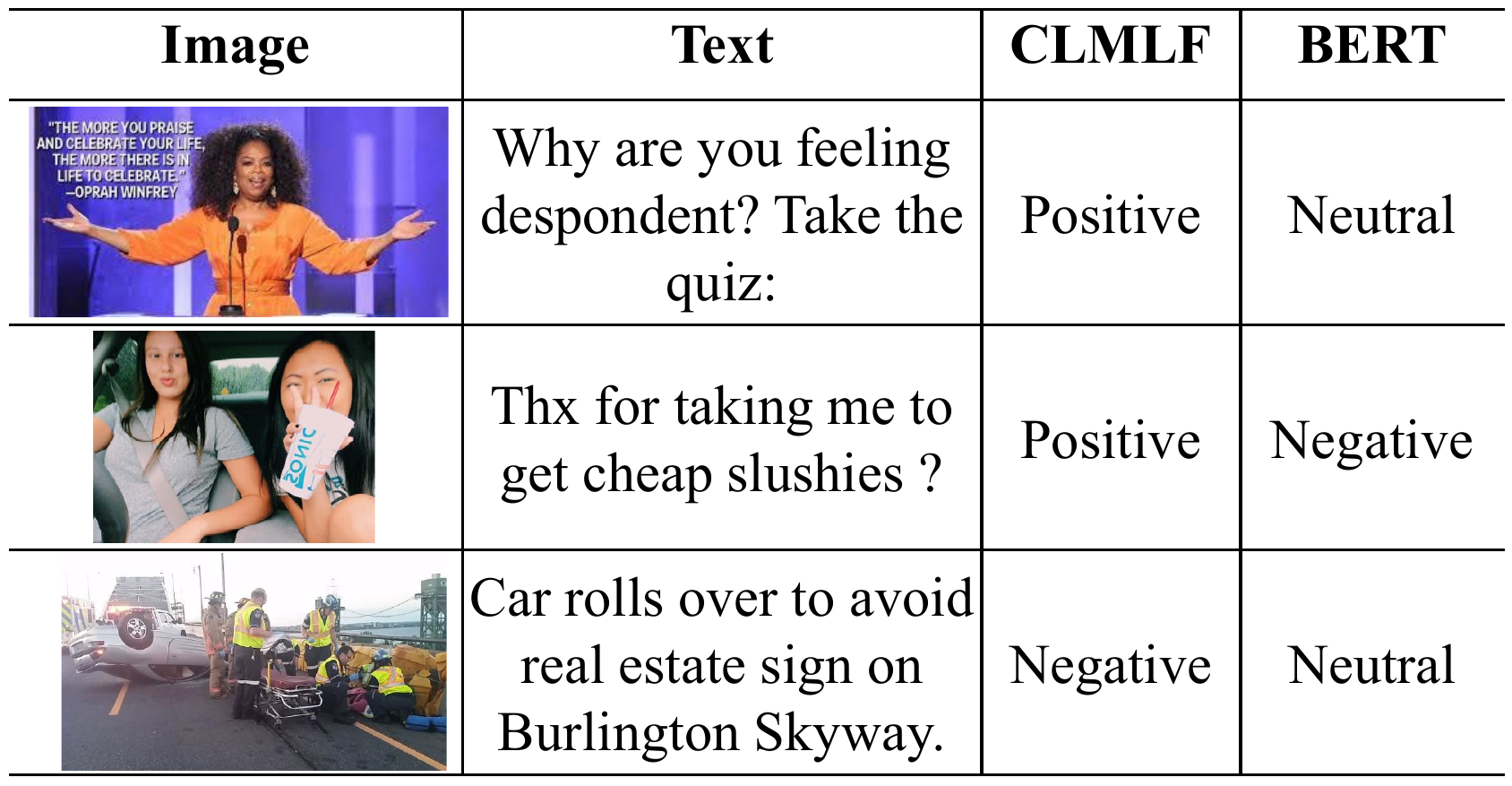}
    \caption{Example of misclassified by BERT and correctly classified by CLMLF}
    \label{fig:case_study}
\end{figure}

To further demonstrate the effectiveness of our model, we give a case study. We compare the sentiment label predicted based on CLMLF and  BERT. As shown in Figure~\ref{fig:case_study}, We can find that if we only consider the sentiment of the text, it is difficult to correctly obtain the user's sentimental tendency. For example, for the first data in Figure~\ref{fig:case_study}, the meaning of the text is to refer to the image, and the image expresses a positive meaning. for the second data, if we only observe the text, we find that it may express negative sentiments. If add the image, we find that it is just a joke and actually expresses positive sentiment. 

\subsection{Visualization}
\label{subsec:visualization}

\begin{figure}[htbp]
\centering
\subfloat[The fishing is a little slow but the \textbf{{\color{red}flowers}} are vibrant and beautiful.]{
\includegraphics[width=0.1\textwidth]{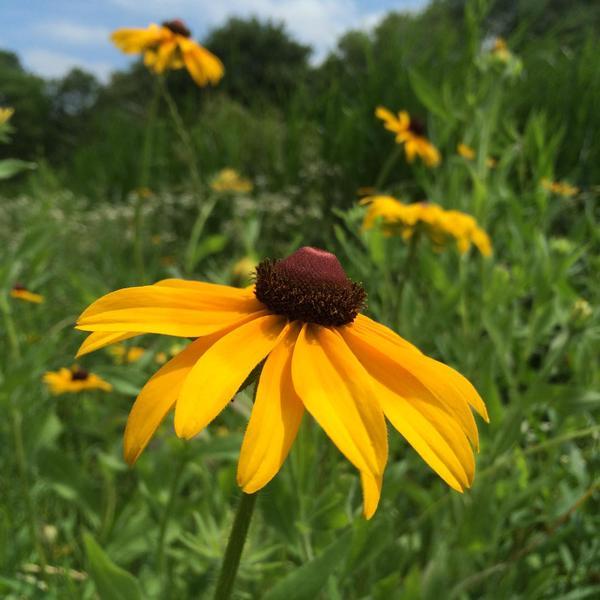}
\includegraphics[width=0.1\textwidth]{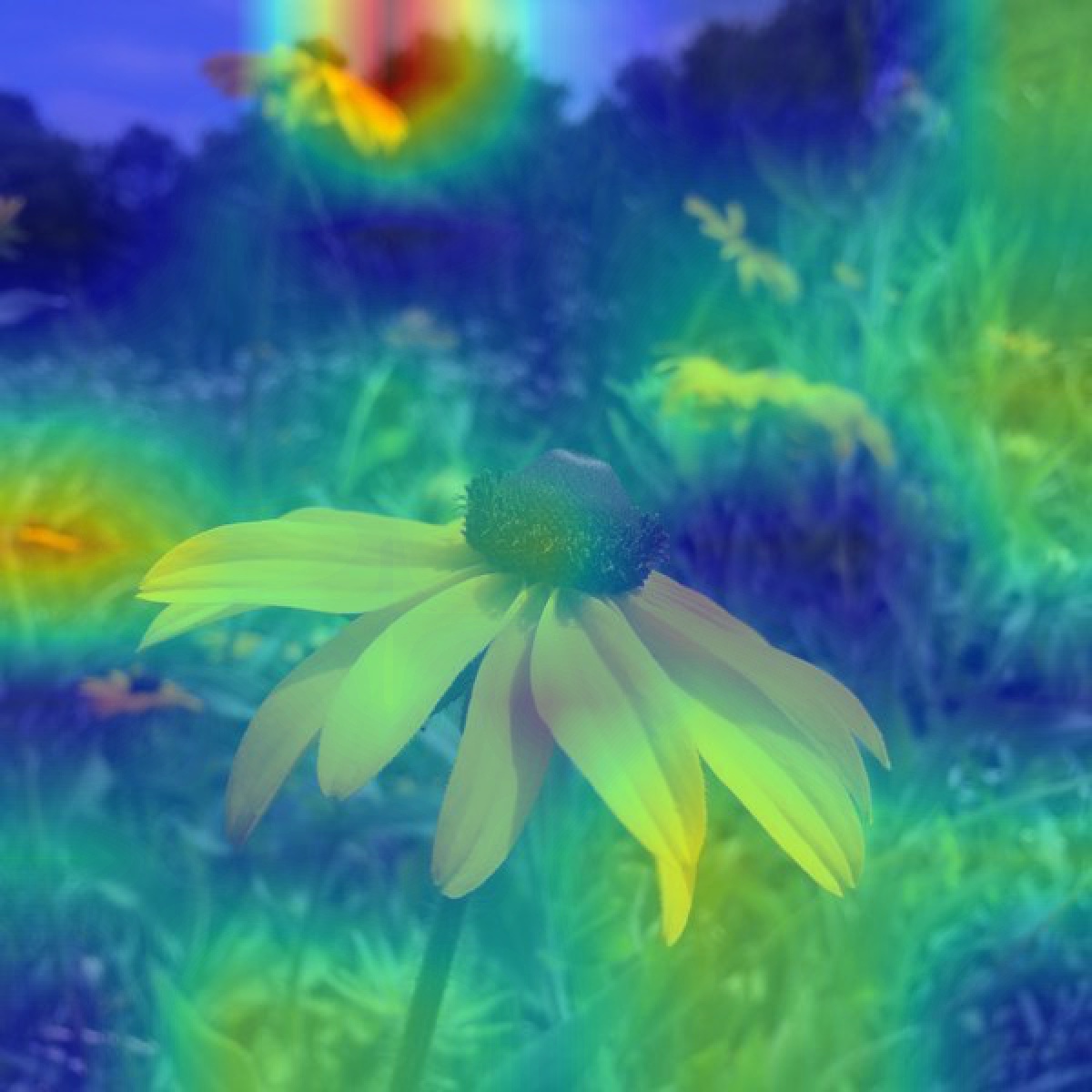}
\label{fig:attention_5}
}
\quad
\subfloat[Kimmy, you're one blessed \textbf{{\color{red}lady}}!]{
\includegraphics[width=0.1\textwidth]{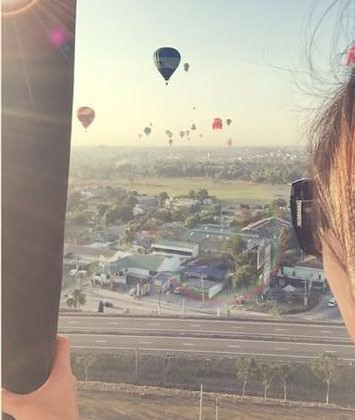}
\includegraphics[width=0.1\textwidth]{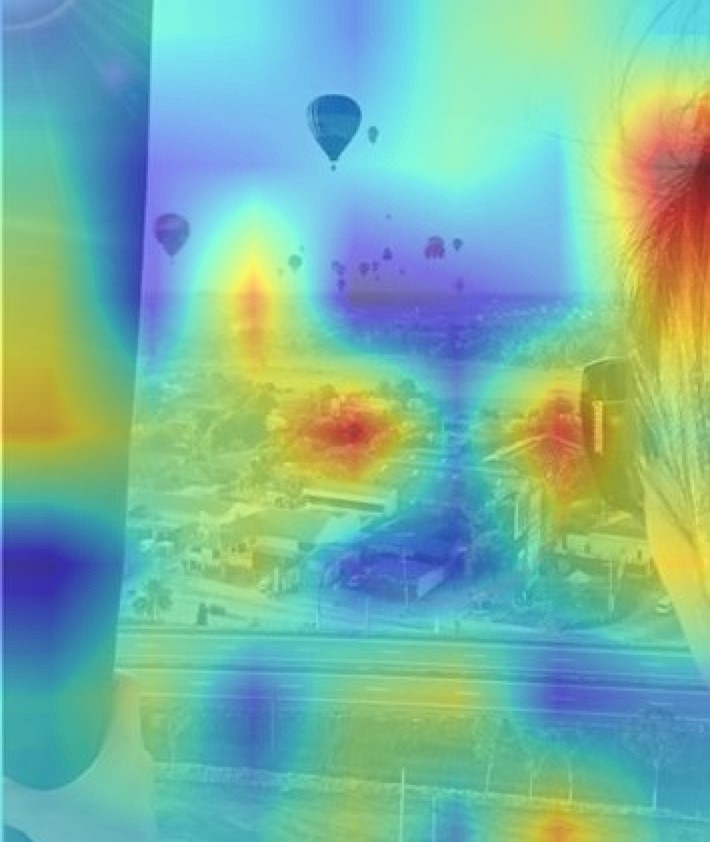}
\label{fig:attention_6}
}
\quad
\subfloat[Martha said for Valentine's Day she wanted a heart shaped \textbf{{\color{red}pancake}} for lunch.]{
\includegraphics[width=0.1\textwidth]{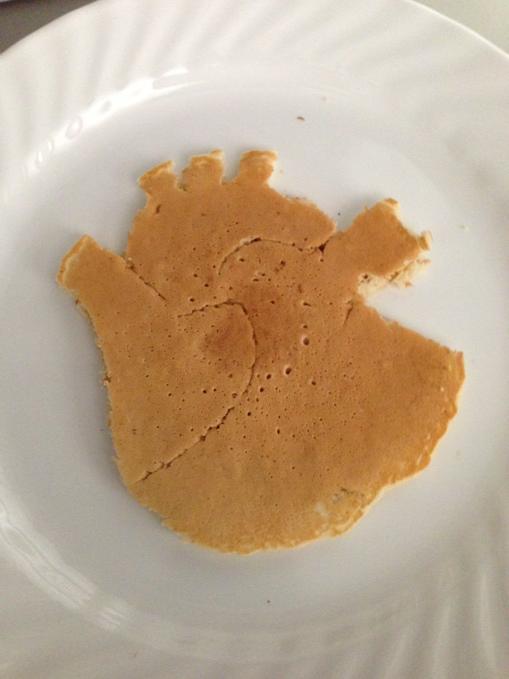}
\includegraphics[width=0.1\textwidth]{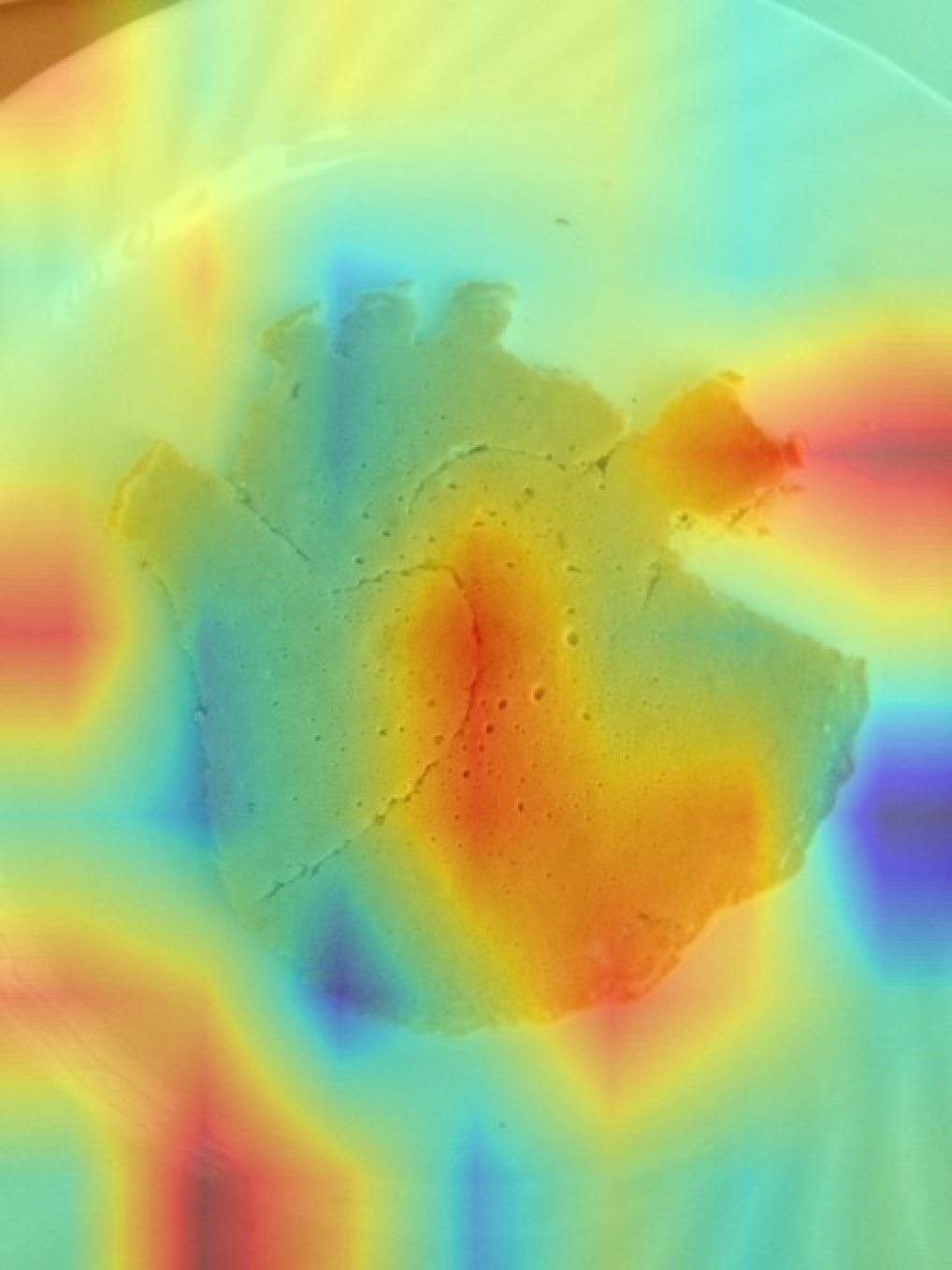}
\label{fig:attention_7}
}
\quad
\subfloat[It is truly a \textbf{{\color{red}hilarious}}, light-hearted read that is a treasure on anyone's bookshelf.]{
\includegraphics[width=0.1\textwidth]{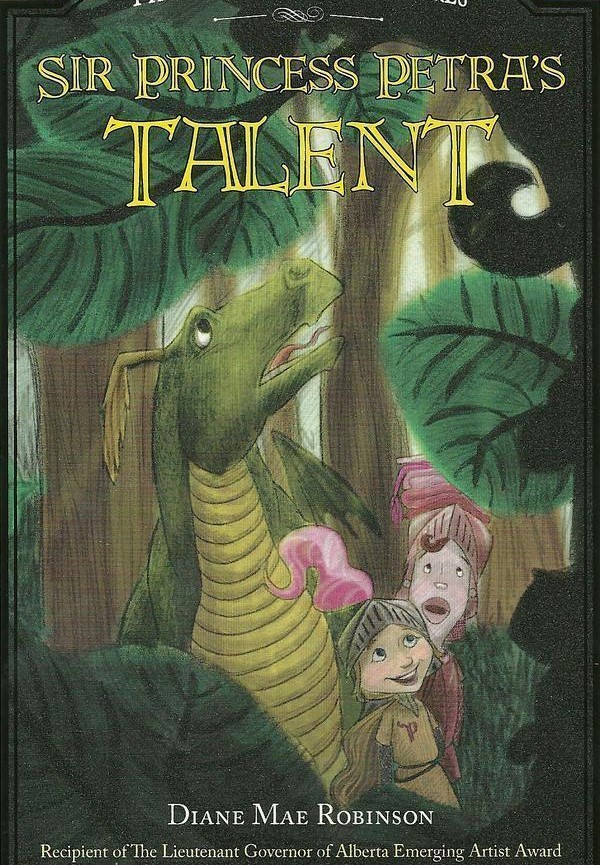}
\includegraphics[width=0.1\textwidth]{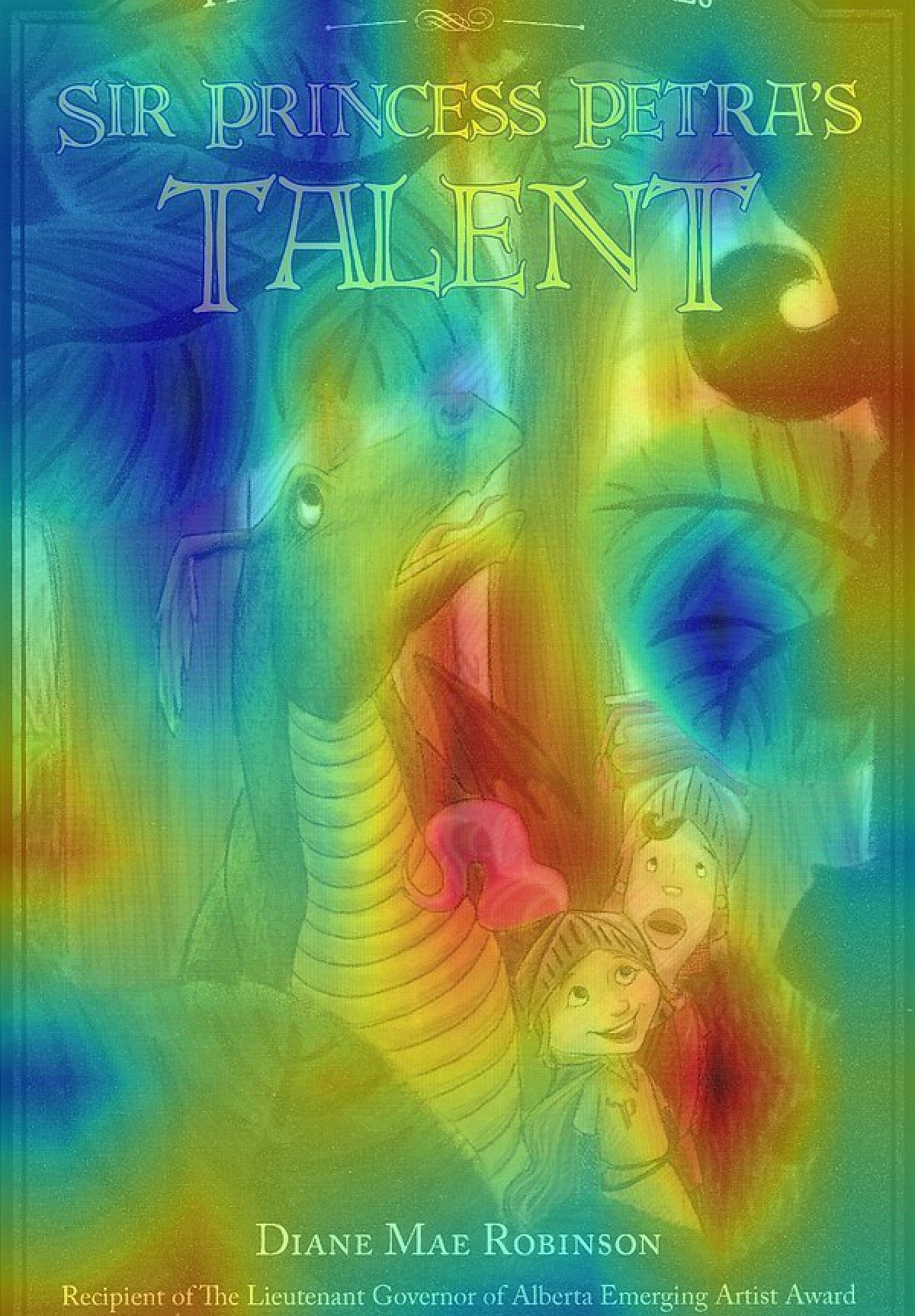}
\label{fig:attention_8}
}
\caption{Attention visualization of some multimodal sentiment data examples}
\label{fig:attention_visualization}
\end{figure}

\textbf{Attention Visualization: } We visualize the attention weight of the first head of the Transformer-Encoder in the last layer of the Multi-Layer Fusion module. The result of the attention visualization is shown in Figure~\ref{fig:attention_visualization}. We can see that for a given keyword, The model can find the target from the image very well and give it more attention weight. This shows that the model aligns the words in the text with the patch area of the image at a token-level, which plays an important role in the model to fuse text and image features. In particular, for Figure~\ref{fig:attention_6}, although "lady" only shows half of the face in the figure, the model still aligns the text and the image very accurately.
These indicate that the model aligns the text and image features at token-level according to our assumptions.

\begin{figure}[htbp]
\centering
\subfloat[BERT]{%
\includegraphics[width=0.21\textwidth]{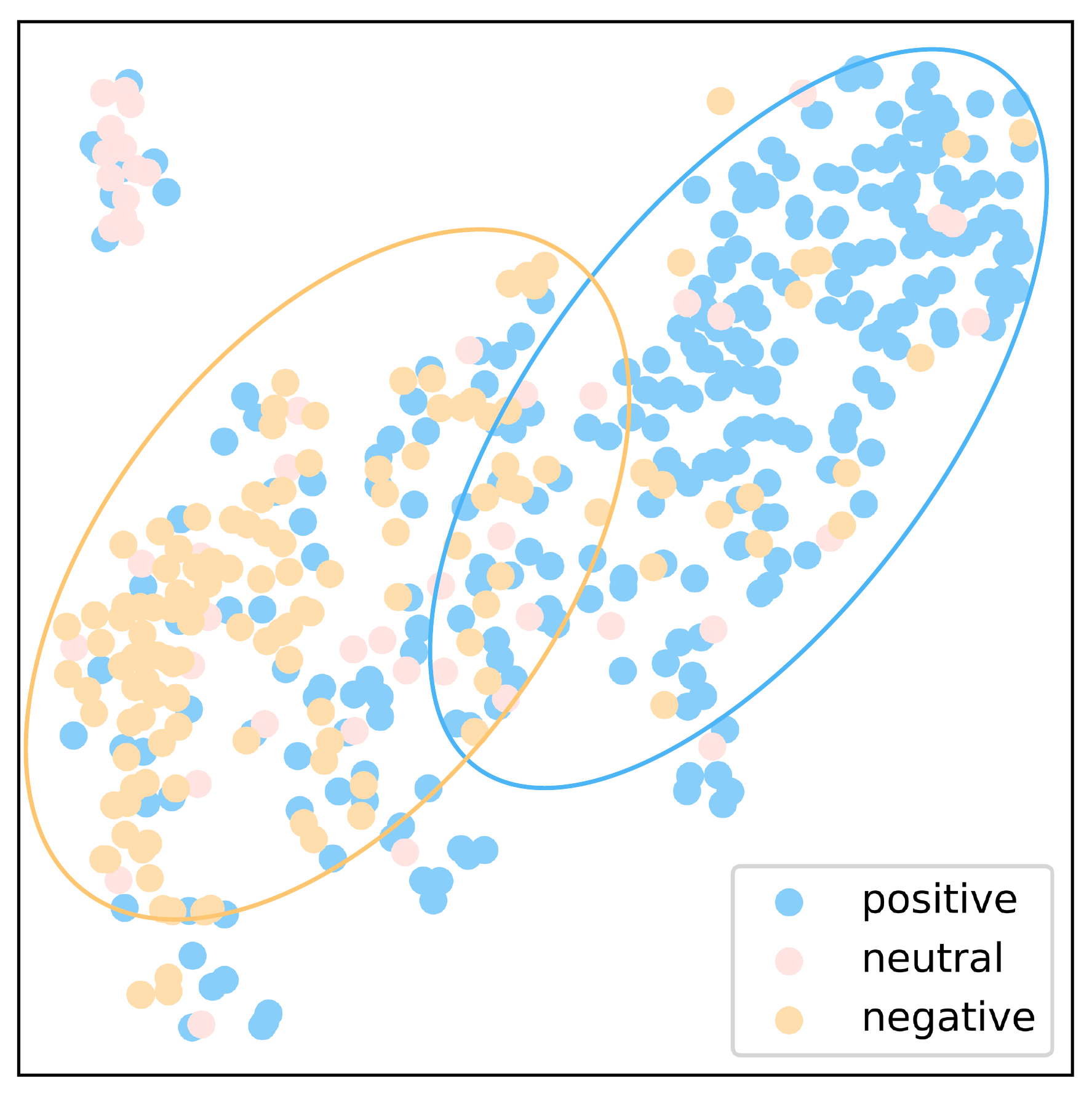}
\label{fig:cluster_bert}
}
\quad
\subfloat[Contrastive Learning]{
\includegraphics[width=0.21\textwidth]{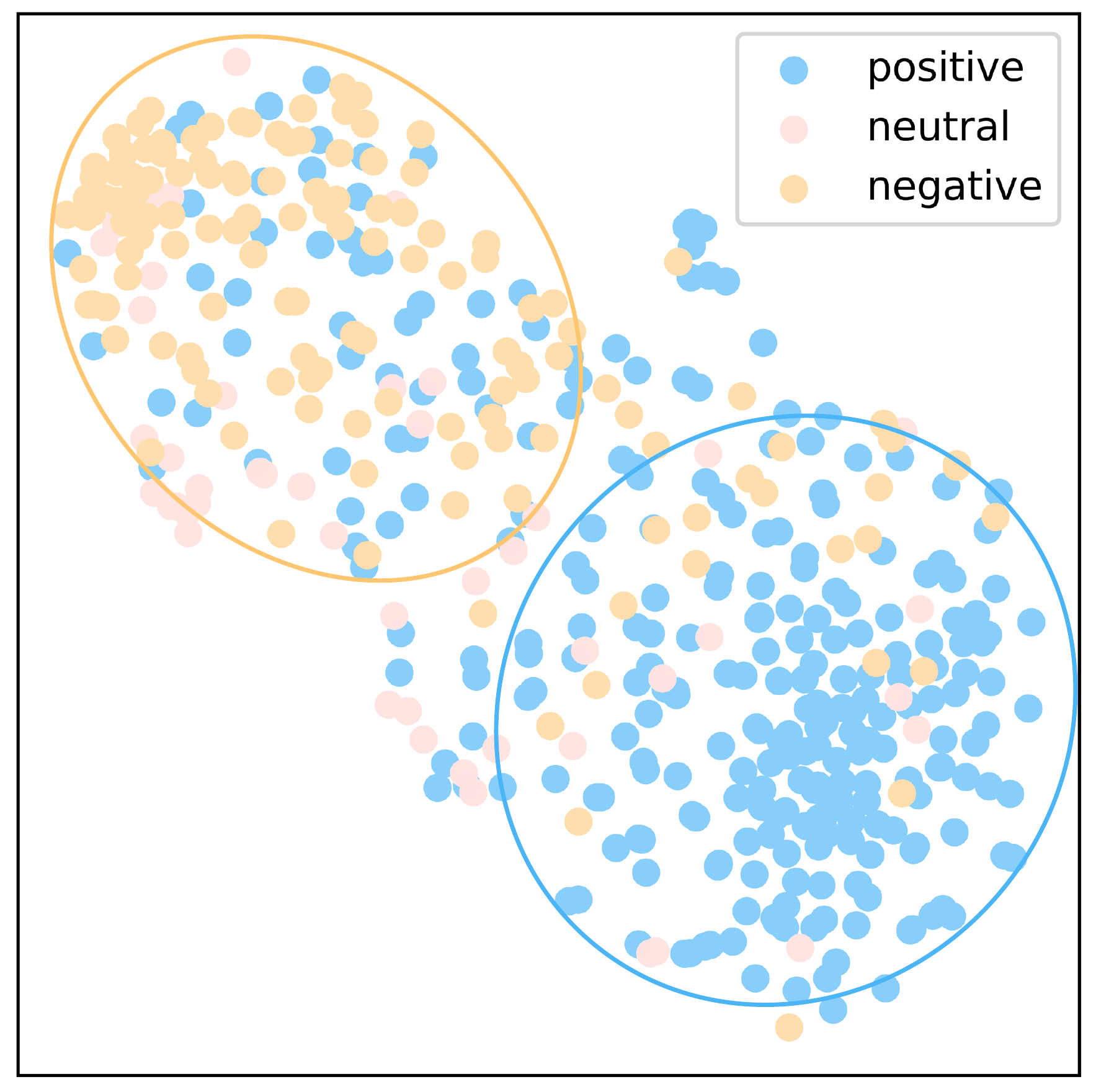}
\label{fig:cluster_with_vl_tf}
}
\caption{Cluster visualization of MVSA-Single}
\label{fig:cluster_visualization}
\end{figure}

\textbf{Cluster Visualization}: In order to verify that our proposed contrastive learning tasks can help the model to learn common features related to sentiment in multimodal data, we conducted a visualization experiment on the MVSA-Single dataset. The data feature vector of the last layer of the model is visualized by dimensionality reduction. We use the TSNE dimensionality reduction algorithm to obtain a 2-dimensional feature vector and visualize it,
as shown in Figure~\ref{fig:cluster_visualization}, Figure~\ref{fig:cluster_bert} is the visualization of the [CLS] of the Bert-base model, and Figure~\ref{fig:cluster_with_vl_tf} shows the visualization of the fusion result output from the CLMLF model. From the figure, we can see that after adding contrastive learning, the distance between positive sentiment and negative sentiment in the vector space is greater, and the degree of data aggregation is more obvious. This shows that the model distinguishes these data in vector space according to common features existing in the same sentimental data. Because the number of neutral sentiment data is relatively small, among the visualization results of the two models, CLMLF's visualization results obviously gather the neutral data together, rather than scattered in the vector space like Bert. All these indicate that adding contrast learning can help the model to learn common features related to sentiment which can improve the performance of the model.

\section{Related Work}

\subsection{Multimodal Sentiment Analysis}

In recent years, deep learning models have achieved promising results for multimodal sentiment analysis. MultiSentiNet~\citep{xu2017multisentinet} and HSAN~\citep{xu2017analyzing} use LSTM and CNN to encode texts and images to get hidden representations, then concatenate texts and images hidden representations to fuse multimodal features. CoMN~\citep{xu2018co} uses a co-memory network to iteratively model the interactions between visual contents and textual words for multimodal sentiment analysis. \citet{yu2019entity} proposes an 
aspect sensitive attention and fusion network to effectively model the intra-modality interactions including aspect-text and aspect-image alignments, and the inter-modality interactions. MVAN~\citep{yang2020image} applies interactive learning of text and image features through the attention memory network module, and the multimodal feature fusion module is constructed by using a multi-layer perceptron and a stacking-pooling module. \citet{yang2021multimodal} uses multi-channel graph neural networks with sentiment-awareness which is built based on the global characteristics of the dataset for multimodal sentiment analysis.

\subsection{Contrastive Learning}

Self-supervised learning attracts many researchers for its soaring performance on representation learning in the last several years~\citep{liu2021self, jing2020self, jaiswal2021survey}. 
% Specifically, It aims at embedding augmented versions of the same sample close to each other while trying to push away embeddings from different samples \cite{jaiswal2021survey}. 
Many models based on contrastive learning have been proposed in both natural language processing and computer vision fields. ConSERT~\citep{yan2021consert}, SimCSE~\citep{gao2021simcse}, CLEAR\citep{wu2020clear} proposed the application of contrastive learning in the field of natural language processing. MoCo~\citep{he2020momentum}, SimCLR~\citep{chen2020simple}, SimSiam~\citep{chen2021exploring}, CLIP~\citep{radford2021learning} proposed the application of contrastive learning in the field of computer vision, and they also have achieved good results in zero-shot learning and few-shot learning. Recently, contrastive learning has been more and more widely used in the field of multimodality.
\citet{huang2021multilingual} uses intra-modal, inter-modal, and cross-lingual contrastive learning which can significantly improves the performance of video search.
\citet{yuan2021multimodal} exploits intrinsic data properties within each modality and semantic information from cross-modal correlation simultaneously, hence improving the quality of learned visual representations.

Compared with the above works, we focus on how to align and fuse the token-level features and learn the common features related to sentiment to further improve the performance of model.

\section{Conclusion and Future Work}

In this paper, we propose a contrastive learning and multi-layer fusion method for multimodal sentiment detection.
Compared with previous works, our proposed MLF module performs multimodal feature fusion from the fine-grained token-level, which is more conducive to the fusion of local features of text and image.
At the same time, we design learning tasks based on contrastive learning to help the model learn sentiment related features in the multimodal data and improve the ability of the model to extract and fuse features of multimodal data.
The experimental results on public datasets demonstrate that our proposed model is competitive with strong baseline models.
Especially through visualization, the contrastive learning tasks and multi-layer fusion module we proposed can be verified with intuitive interpretations.
In future work, we will incorporate other modalities such as audio into the sentiment detection task.

\section*{Acknowledgements}

We are grateful to the anonymous reviewers, area chair and Program Committee for their insightful comments and suggestions.
The work of this paper is funded by the project of National
key research and development program of China (No.
2020YFB1406902).

% \newpage
% \clearpage

\bibliography{acl_latex}
\bibliographystyle{acl_natbib}

\clearpage
% \newpage

\appendix

\section{Dataset Statistics}
\label{sec:dataset_detailed_statistics}

The detailed statistics for the MVSA-Single, MVSA-Multiple and HFM datasets are listed in Table~\ref{tab:dataset_detailed_statistics}. We can see that HFM is a binary classification multimodal sentiment dataset, while MVSA-Single and MVSA-Multiple are three classification multimodal sentiment datasets.

\begin{table}[htb]
\centering
\begin{tabular}{c|c|c|c|c}
\hline
 \textbf{Dataset} & \textbf{Label} & \textbf{Train} & \textbf{Val} & \textbf{Test}  \\
\hline
 \multirow{3}{*}{\makecell[c]{MVSA-\\Single}} & Positive & 2147 & 268 & 268 \\
 & Neutral & 376 & 47 & 47  \\
 & Negative & 1088 & 135 & 135  \\
\hline
 \multirow{3}{*}{\makecell[c]{MVSA-\\Multiple}} & Positive & 9056 & 1131 & 1131   \\
 & Neutral & 3528 & 440 & 440  \\
 & Negative & 1040 & 129 & 129  \\
\hline
 \multirow{2}{*}{HFM} & Positive & 8642 & 959 & 959  \\
 & Negative & 11174 & 1451 & 1450  \\
\hline
\end{tabular}

\caption{Number of data for each sentiment category in each dataset}
\label{tab:dataset_detailed_statistics}
\end{table}

\section{Aspect Based Multimodal Sentiment}
\label{sec:aspect_based_multimodal_sentiment}

\subsection{Experimental Setup}

Because CLMLF is designed for sentence-level multimodal sentiment analysis, we have made some minor changes to the input of CLMLF model to adapt to aspect based multimodal sentiment analysis. We change the input form from "[CLS] sentence [SEP]" to "[CLS] sentence [SEP] aspect [SEP]" and no change the input of image modality. Although this change is very simple, CLMLF can work well in aspect based multimodal sentiment analysis tasks and achieves good results.

We use three aspect based multimodal sentiment dataset: Multi-ZOL\footnote{https://github.com/xunan0812/MIMN}~\citep{xu2019multi}, Twitter-15~\citep{zhang2018adaptive}  and Twitter-17\footnote{https://github.com/jefferyYu/TomBERT}~\citep{lu2018visual}. The statistics of these datasets are given in Table~\ref{tab:aspect_data_statistics}. Compared with the dataset of sentence-level multimodal sentiment analysis, each sentence will have a corresponding aspect attribute. Especially for the Multi-ZOL dataset, each data contains multiple images. And we only randomly select one image for fusion. Although some features are lost, the experimental results show that it is improved compared with the only text modality.

\begin{table}[htb]
\centering
\begin{tabular}{c|cccc}
\hline
\textbf{Dataset} & \textbf{Train} &  \textbf{Val} & \textbf{Test} & \textbf{Total} \\
\hline
Multi-ZOL & 22743 & 2843 & 2843 & 28429 \\
Twitter-15 & 3179 & 1122 & 1037 & 5338 \\
Twitter-17 & 3562 & 1176 & 1234 & 5972 \\
\hline
\end{tabular}

\caption{Statistics of the three datasets}
\label{tab:aspect_data_statistics}
\end{table}

\begin{table*}[!t]
\centering
\begin{tabular}{c|c|cc|cc|cc}
\hline
\multirow{2}{*}{\textbf{Modality}} & \multirow{2}{*}{\textbf{Model}} & \multicolumn{2}{c|}{\textbf{Multi-ZOL}} & \multicolumn{2}{c|}{\textbf{Twitter-15}} &  \multicolumn{2}{c}{\textbf{Twitter-17}} \\
& & Acc & F1 & Acc & F1 & Acc & F1 \\
\hline
\multirow{3}{*}{\textbf{Text}} & LSTM & 0.5892 & 0.5729 & 0.6798 & 0.5730 & 0.5592 & 0.5169  \\
& AE-LSTM & 0.5958 & 0.5895 & 0.7030 & 0.6343 & 0.6167 & 0.5797 \\
& RAM & 0.6018 & 0.5968 & 0.7068 & 0.6305 & 0.6442 & 0.6101 \\
& BERT & 0.6959 & 0.6868 & 0.7387 & 0.7023 & 0.6848 & 0.6553 \\
\hline
\multirow{5}{*}{\textbf{Multimodal}} & MIMN & 0.6159 & 0.6051 & 0.7184 & 0.6569 & 0.6588 & 0.6299 \\
& ESAFN & - & - & 0.7338 & 0.6737 & 0.6783 & 0.6422 \\   
& TomBERT & - & - & 0.7715 & 0.7175 & 0.7034 & 0.6803 \\   
& CLMLF & \textbf{0.7452} & \textbf{0.7075} & \textbf{0.7811} & \textbf{0.7460}  & \textbf{0.7098} & \textbf{0.6913} \\  
\hline
\end{tabular}

\caption{Experimental results of different models on aspect based datasets}
\label{tab:aspect_result}
\end{table*}

\begin{table*}[htbp]
\centering
\begin{tabular}{ccc|cc|cc|cc}
\hline
 & \multirow{2}{*}{\textbf{Model}} &&  \multicolumn{2}{c|}{\textbf{Multi-ZOL}} & \multicolumn{2}{c|}{\textbf{Twitter-15}} &  \multicolumn{2}{c}{\textbf{Twitter-17}} \\
 &&& Acc & F1 & Acc & F1 & Acc & F1 \\
\hline
 & BERT && 0.6959 & 0.6868 & 0.7387 & 0.7023 & 0.6848 & 0.6553 \\
 & +MLF && 0.7301 & 0.6897 & 0.7424 & 0.7017 & 0.6848 & 0.6579  \\
 & +MLF, LBCL && 0.7336 & 0.6953 & 0.7715 & 0.7311 & 0.6969 & 0.6790  \\
 & +MLF, DBCL && 0.7347 & 0.7015 & 0.7445 & 0.6964 & 0.6921 & 0.6722  \\
\hline
& CLMLF && \textbf{0.7452} & \textbf{0.7075} & \textbf{0.7811} & \textbf{0.7460}  & \textbf{0.7098} & \textbf{0.6913} \\  
\hline
\end{tabular}

\caption{Ablation results of CLMLF}
\label{tab:aspect_ablation}
\end{table*}

\subsection{Results}

We compare our model with other baseline models:

\begin{itemize}
\item LSTM, a standard sentence-level LSTM model without explicitly considering the aspect. Therefore, this result is also the worst.

\item AE-LSTM~\citep{wang2016attention}, an attention-based LSTM for aspect-level sentiment classification, which uses the attention mechanism to capture the important context information related to the aspect.

\item RAM~\citep{chen2017recurrent}  is a memory based model, which builds memory on the hidden states of a Bi-LSTM and generates aspect representation based on a Bi-LSTM. Then pays multiple attentions on the memory to pick up important information to predict the final sentiment, by combining the features from different attentions non-linearly.

\item MIMN~\citep{xu2019multi}, the multimodal approach for aspect-level sentiment classification task, which adopts multi-hop memory network to model the interactive attention between the aspect word, the textual context, and the visual context.

\item TomBERT~\citep{yu2019adapting}, a multimodal model which  borrow the idea from self-attention and design a target attention mechanism to perform target-image matching to derive target sensitive visual representations.

\item ESAFN~\citep{yu2019entity}  proposes an entity-sensitive attention and fusion network which capture the intra-modality dynamics by leverages an effective attention mechanism to generate entity-sensitive textual and visual representations. And uses visual attention mechanism to learn the entity-sensitive visual representation. Moreover, ESAFN further fuses the textual and visual representations with a bilinear interaction layer.

\end{itemize}

Table~\ref{tab:aspect_result} illustrates the performance comparison of our CLMLF mdoel with the baseline methods. We use Macro-F1 and ACC as the evaluation metrics for all datasets. The experimental results show that CLMLF can still achieve good results. We also conducted ablation experiments, as shown in Table~\ref{tab:aspect_ablation}. The experiments again proved that the multi-layer fusion module, label based contrastive learning task and data based contrastive task we proposed are effective.

\end{document}